\documentclass[letterpaper]{article} % DO NOT CHANGE THIS
\usepackage{aaai25}  % DO NOT CHANGE THIS
\usepackage{times}  % DO NOT CHANGE THIS
\usepackage{helvet}  % DO NOT CHANGE THIS
\usepackage{courier}  % DO NOT CHANGE THIS
\usepackage[hyphens]{url}  % DO NOT CHANGE THIS
\usepackage{graphicx} % DO NOT CHANGE THIS
\usepackage{natbib}  % DO NOT CHANGE THIS AND DO NOT ADD ANY OPTIONS TO IT
\usepackage{caption} % DO NOT CHANGE THIS AND DO NOT ADD ANY OPTIONS TO IT
\usepackage{algorithm}
\usepackage{algorithmic}

\usepackage{times}
\usepackage{multirow}
\usepackage{amsfonts}
\usepackage{subfig}
\usepackage[table]{xcolor}
\usepackage{colortbl}
\usepackage{cite}
\usepackage{diagbox}
\usepackage{rotating}
\usepackage{booktabs}
\usepackage{overpic}
\usepackage{contour}
\usepackage{bbding}
\usepackage{times}
\usepackage{epsfig}
\usepackage{graphicx}
\usepackage{amsmath}
\usepackage{amssymb}
\usepackage{newfloat}
\usepackage{listings}
\DeclareCaptionStyle{ruled}{labelfont=normalfont,labelsep=colon,strut=off} % DO NOT CHANGE THIS
\floatstyle{ruled}
\newfloat{listing}{tb}{lst}{}
\floatname{listing}{Listing}
\title{Generative Planning with 3D-vision Language Pre-training for End-to-End Autonomous Driving}
\author{
   Tengpeng Li\textsuperscript{\rm 1,2}, Hanli Wang\textsuperscript{\rm 1,2}\thanks{Corresponding authors.}, Xianfei Li\textsuperscript{\rm 3}, Wenlong Liao\textsuperscript{\rm 3}, Tao He\textsuperscript{\rm 3,4}, Pai Peng\textsuperscript{\rm 3}\footnotemark[1]\\
}

\affiliations{
    \textsuperscript{\rm 1}College of Electronic and Information Engineering, Tongji University, Shanghai, China\\
    \textsuperscript{\rm 2}School of Computer Science and Technology, Tongji University, Shanghai, China\\
    \textsuperscript{\rm 3}COWAROBOT, China\\
    \textsuperscript{\rm 4}School of Electronic Engineering, University of South China, Hunan, China\\
    hanliwang@tongji.edu.cn, pengpai\_sh@163.com
}

\usepackage{bibentry}
\begin{document}

\maketitle

\begin{abstract}
Autonomous driving is a challenging task that requires perceiving and understanding the surrounding environment for safe trajectory planning. While existing vision-based end-to-end models have achieved promising results, these methods are still facing the challenges of vision understanding, decision reasoning and scene generalization. To solve these issues, a generative planning with 3D-vision language pre-training model named GPVL is proposed for end-to-end autonomous driving. The proposed paradigm has two significant aspects. On one hand, a 3D-vision language pre-training module is designed to bridge the gap between visual perception and linguistic understanding in the bird's eye view. On the other hand, a cross-modal language model is introduced to generate holistic driving decisions and fine-grained trajectories with perception and navigation information in an auto-regressive manner. Experiments on the challenging nuScenes dataset demonstrate that the proposed scheme achieves excellent performances compared with state-of-the-art methods. Besides, the proposed GPVL presents strong generalization ability and real-time potential when handling high-level commands in various scenarios. It is believed that the effective, robust and efficient performance of GPVL is crucial for the practical application of future autonomous driving systems. Code is available at \url{https://github.com/ltp1995/GPVL}
\end{abstract}

%\texttt{https://github.com/ltp1995/GPVL}

\begin{figure}[h]
\begin{center}
\includegraphics[width=0.95\linewidth]{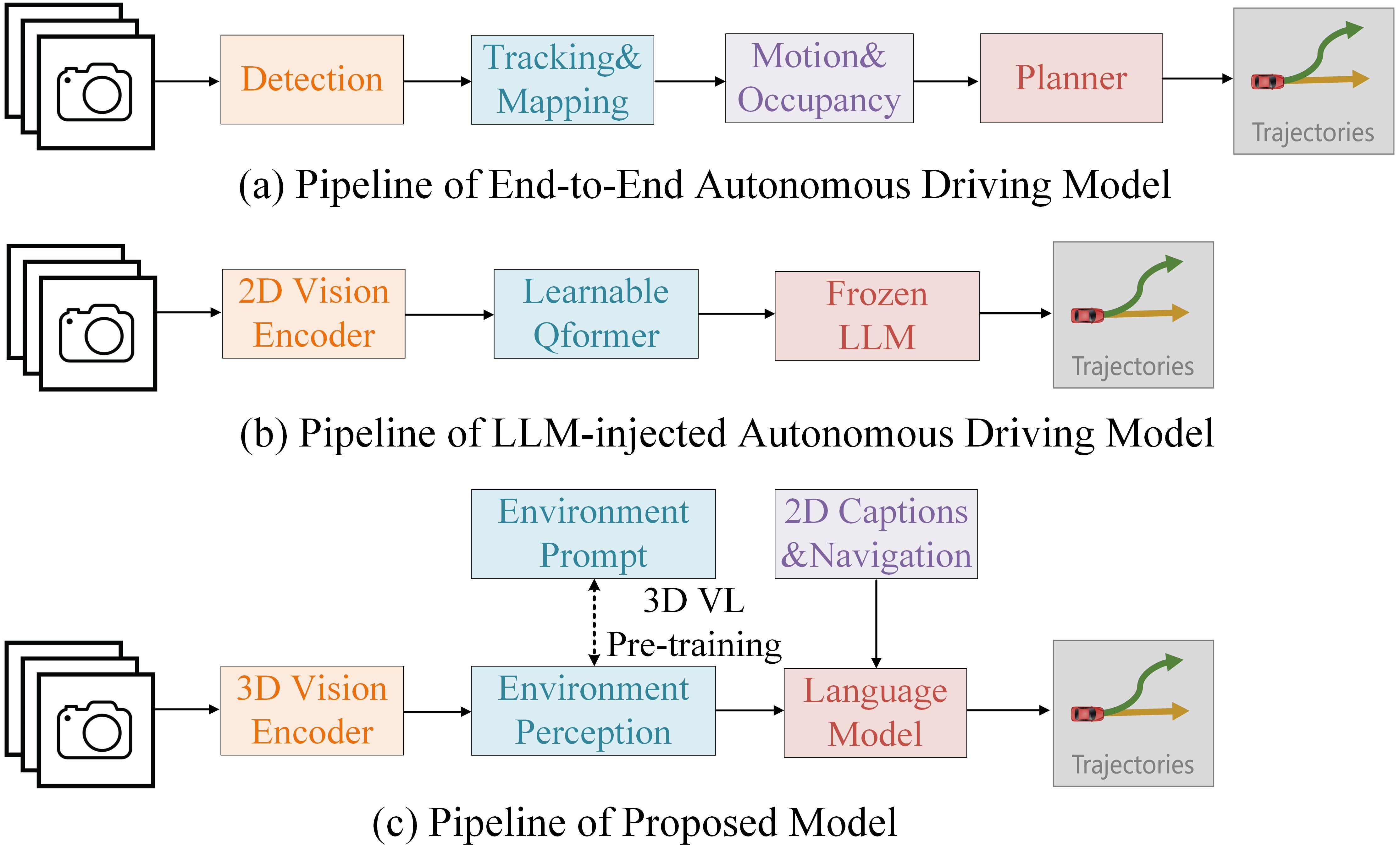}
\end{center}
\caption{(a) The existing end-to-end autonomous driving framework only utilizes visual information to complete perception, prediction and planning tasks. (b) The emerging LLM-injected autonomous driving models merely introduce 2D visual features and use time-consuming LLM for planning decision. (c) The designed scheme focuses on 3D-vision language pre-training and conducts the planning by a language generation style.}
\label{fig:paradigm}
\end{figure}

\section{Introduction}
\label{sec:introduction}

Autonomous driving is a challenging task that requires a profound understanding about the surrounding environment of autonomous vehicles to ensure safety and efficient real-world deployment. An excellent autonomous driving system must have the capacity to comprehensively perceive the driving environment and precisely make the planning decision based on the road information.

Recently, several end-to-end autonomous driving frameworks are proposed and achieve promising results, by understanding the driving scenes with sensor data and outputting planning decisions. Some early methods~\cite{codevilla2019exploring, prakash2021multi} directly obtain the predicted planning trajectories without thorough understanding of driving scenes, resulting in limited interpretability and difficulty in convergence. As illustrated in Fig.~\ref{fig:paradigm}(a), most schemes~\cite{hu2022st,hu2023planning,jiang2023vad} utilize the sensor information to integrate multiple vision tasks (\emph{e.g.}, 3D object detection and motion prediction) in one network to enhance planning performance. However, these vision-only approaches struggle to incorporate contextual cues for safe decision-making and require further optimization. With the rapid development of large language models~(LLM), the outstanding reasoning ability has attracted much interest and promoted applications in the autonomous driving area. As presented in Fig.~\ref{fig:paradigm}(b), several recent methods~\cite{xu2023drivegpt4,wang2023drivemlm} introduce multi-modal large language models~(MLLM) to produce driving explanations and language-injected planning results. Nonetheless, these models pre-trained on 2D vision-language datasets struggle with understanding the complex 3D spatial relations in driving scenarios. Therefore, several effective strategies are designed to overcome above challenges, such as directly using the detected 3D bounding boxes of crucial objects~\cite{tian2024drivevlm} and integrating the bird's eye view~(BEV) feature with learnable queries~\cite{ding2024holistic}. However, these advanced strategies merely integrate 2D and 3D visual information without adequately addressing the alignment of 3D visual features with linguistic representations.

To address the aforementioned challenges, a novel generative planning with 3D-vision language pre-training~(GPVL) paradigm is proposed for autonomous driving, as shown in Fig.~\ref{fig:paradigm}(c). GPVL first utilizes the pre-trained BEVformer~\cite{li2022bevformer} to extract the BEV feature map of multi-view images, which covers the essential semantic elements~(\emph{e.g.}, vehicles, pedestrians and traffic lines) of driving scenes. Motivated by VAD~\cite{jiang2023vad}, the vectorized detection, motion and mapping transformers are introduced to learn the crucial perception information. Then, a 3D-vision language pre-training module is developed to align BEV features with linguistic representations, enabling comprehensive 3D scene understanding and textual reasoning in a shared feature space, where the group-wise alignment establishes the multi-level associations between vision and language features. Furthermore, GPVL designs a 2D scene captioning model based on the pretrained BLIP~\cite{li2022blip}, which is trained and tested on the Ominidrive-nuScenes dataset~\cite{wang2024omnidrive} to generate scene-level descriptions. Finally, the 2D visual captioning, the aligned 3D perception feature and navigation instruction are fed into a language model to generate the holistic driving decision and fine-grained trajectory in an auto-regressive manner.

The major contributions of this work are summarized in the following three aspects.
\begin{itemize}
\item
A 3D-vision language pre-training module is presented to establish group-wise correlations between vision and language features, facilitating a thorough understanding of the driving environment.
\item
A cross-modal language model is developed to generate decisions and trajectories using captioning, perception and navigation information in an auto-regressive manner, endowing the model with reasoning and generation capabilities.
\item
We propose a generative planning with 3D-vision language pre-training framework, which learns language-guided perception features and generates contextual trajectories, thereby enhancing system safety.
\end{itemize}

\section{Related Work}
\subsection{End-to-End Autonomous Driving}

The goal of end-to-end autonomous driving is to construct a completely differentiable system that can convert sensor data into vehicle control
commands~\cite{prakash2021multi,wu2022trajectory}. This system integrates all modules (\emph{i.e.}, perception, prediction, planning and control) for interpretable end-to-end optimization, which can reduce the accumulated errors of all modules and enhance the safety of the autonomous driving system. Recently, several effective models are proposed for this challenging task and can be categorized into open-loop autonomous driving methods and closed-loop autonomous driving methods. Many schemes~\cite{hu2023planning, gu2023vip3d, jiang2023vad, pan2024vlp} address this task in the open-loop manner, whose training and evaluation stages are conducted on real-world datasets. For example, Hu~\textit{et al.}~\cite{hu2023planning} propose an innovative planning-oriented approach to interactively optimize all blocks. Jiang~\textit{et al.}~\cite{jiang2023vad} develop a vectorized scene representation technology to meet real-time requirements. In contrast, closed-loop methods use simulators like CARLA~\cite{dosovitskiy2017carla} to create virtual environments for the feedback of other agents on the road. Specifically, Jia~\textit{et al.}~\cite{jia2023think} optimize future position and action features twice to improve model safety. Chen~\textit{et al.}~\cite{chen2024vadv2} utilize the environmental features to learn the probabilistic distribution of actions. Nevertheless, these methods have difficulty to generate robust trajectories in a reasonable way for efficient vehicle control. In this work, the proposed GPVL is capable of performing reasonable analysis and generating contextual trajectory.

\subsection{Large Language Models for Autonomous Driving}

Recently, many researches~\cite{driess2023palm, wang2023voyager, brohan2023rt} employ LLMs to endow autonomous driving systems with the embodied intelligence capability and achieve favorable results in the decision stage. Inspired by the success in the field of robotics, several works~\cite{wen2023dilu,mao2023gpt} attempt to use LLMs for autonomous driving. In particular, Dilu~\cite{wen2023dilu} and GPT-driver~\cite{mao2023gpt} utilize LLM to generate reasonable trajectories by regarding the perception information of agents as language prompts. Furthermore, recent MLLMs~\cite{driess2023palm, li2023blip2, bai2023qwen, liu2023visual} introduce both visual encoder and LLM to successfully establish the semantic correlations between vision and language, which have attracted much interest in the autonomous driving area. For example, Shao~\textit{et al.}~\cite{shao2023lmdrive} learn an integrated representation of multi-modal sensor data and human navigation instruction to output reliable control commands. Ding~\textit{et al.}~\cite{ding2024holistic} incorporate the BEV representation into multi-view video features to enhance the understanding of driving scenes. However, the fused multi-modal features of these schemes either lack 3D spatial awareness of all targets or fail to align with linguistic representations. In this work, the introduced BEV-level visual transformer and the 3D-vision language pre-training module are devised to address these challenges.

\begin{figure*}[htbp]
\begin{center}
\includegraphics[width=0.95\linewidth]{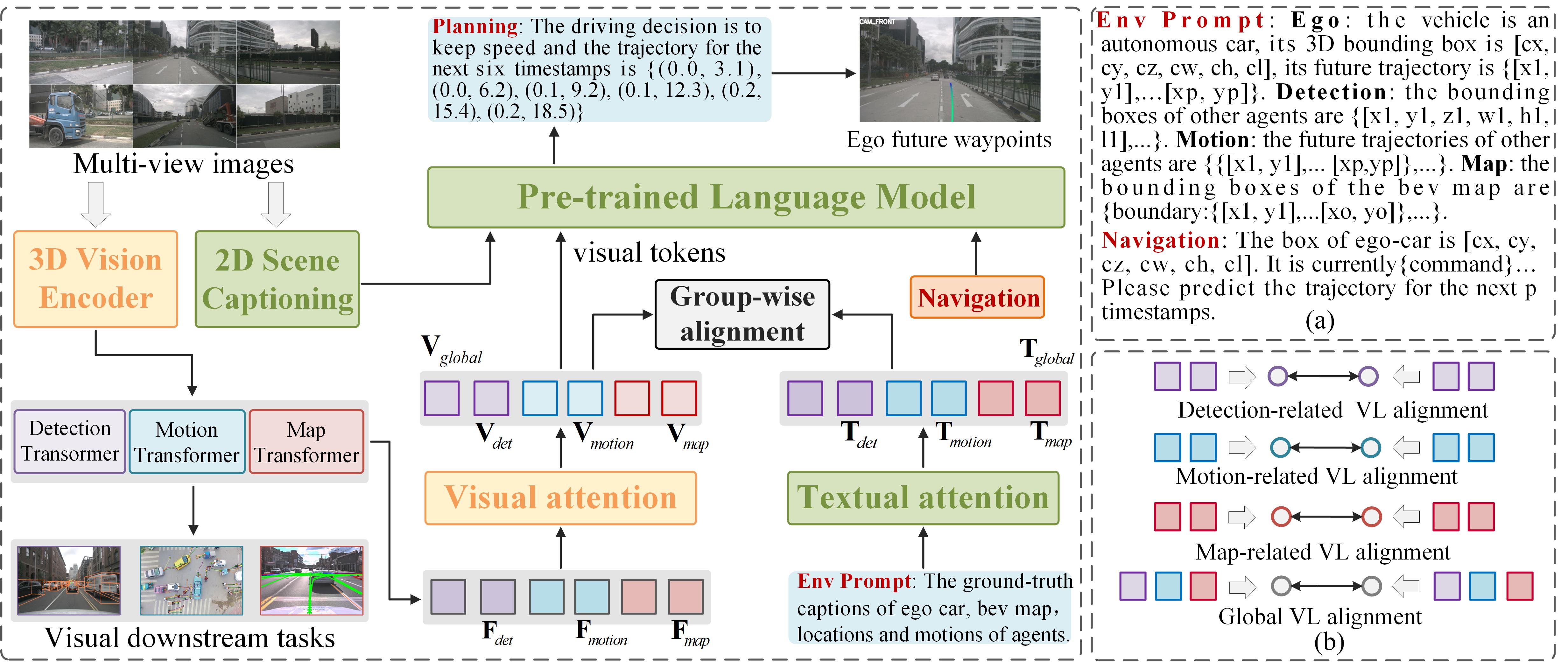}
\end{center}
\caption{Pipeline of GPVL for autonomous driving. The framework is divided into three parts: (1) the backbone includes a 3D-vision encoder to obtain the basic BEV feature, then it is decoded into constrained detection, motion and map features, (2) the 3D-vision language pre-training module establishes the associations between vision and language features with the group-wise alignment, (3) the cross-modal language model generates the future planning decision in an auto-regressive manner based on aligned visual feature and navigation prompt. Note that in (a), the env prompt represents the detailed descriptions of 3D bounding boxes, agent motions and BEV map on the road, and the navigation means the detailed description of high-level instruction for the self-driving car. Four kinds of 3D-vision language~(VL) alignments are presented in (b) for explanation.}
\label{fig:pipeline-overall}
\end{figure*}

\section{Proposed Method}
\label{sect:prop}

The overall framework of the proposed GPVL is presented in Fig.~\ref{fig:pipeline-overall}, including three key components: (1) the backbone generates the supervised detection, map and motion features based on the extracted BEV feature, (2) the 3D-vision language pre-training module aims to align the visual and linguistic features in a common semantic space, and (3) the cross-modal language model produces the reliable decision and trajectory in a generative fashion.

\subsection{Method Overview}
\label{sect:prop:ov}

Suppose $\mathcal{X}=\{{\textbf{X}}_n\}_{n=1}^{N}$ denotes the sampled $N$ multi-view images, the goal of the proposed scheme is to generate the safe driving trajectory of ego vehicle. To acquire the satisfying planning result, GPVL is proposed to establish semantic associations between 3D visual features and linguistic representations, and generates the high-quality route in an auto-regressive manner. Figure~\ref{fig:pipeline-overall} illustrates the detailed structure of the proposed model. Specifically, for the input multi-view images $\mathcal{X}$, we first use the BEV encoder in the pre-trained BEVformer~\cite{li2022bevformer} to extract the BEV feature $\textbf{B}=\mathcal{F}_{bev}(\mathcal{X})$. Afterwards, $\textbf{B}$ is fed into the detection, motion and map transformers to simultaneously learn the 3D object boxes, agent motions and map elements, resulting in constrained detection feature $\textbf{F}_{det}$, motion feature $\textbf{F}_{motion}$ and map feature $\textbf{F}_{map}$. Then, $\textbf{F}_{det}$, $\textbf{F}_{motion}$ and $\textbf{F}_{map}$ are sent into a visual attention block $\mathcal{F}_{vatt}$ composed of multiple transformer layers, generating the attentive visual features $\textbf{V}_{det}$, $\textbf{V}_{motion}$, $\textbf{V}_{map}$ and their concatenation $\textbf{V}_{global}$. Meanwhile, the environment ground-truth descriptions are integrated into a textual attention block $\mathcal{F}_{tatt}$ based on the BERT~\cite{devlin2018bert} structure to obtain textual representations $\textbf{T}_{det}$, $\textbf{T}_{motion}$, $\textbf{T}_{map}$ and their concatenation $\textbf{T}_{global}$. A group-wise alignment module is further designed to project these vision and language features in a shared semantic space. Finally, $\textbf{V}_{global}$ together with the navigation prompt of ego vehicle are sent into the language model to generate the dependable trajectory in a natural language format.

\subsection{3D-vision Language Pre-training}
\label{sect:prop:vlp}

The 3D-vision language pre-training module is developed to establish the multi-level alignment between vision and language modalities through contrastive learning~\cite{radford2021learning}. Several previous methods~\cite{shao2023lmdrive,wang2023drivemlm, ding2024holistic} have introduced pre-trained LLMs~(\emph{e.g.}, LLaMA~\cite{touvron2023llama}) to enhance perception and understanding of driving scenes. However, these approaches either lack 3D spatial information or exhibit semantic gaps between different representations, which hinder accurate target localization and trajectory inference based on extracted features. In contrast, the proposed model utilizes the supervised detection, motion and map features to perceive the 3D spatial distributions of targets and minimizes the semantic gap between different features through the 3D-vision language pre-training module.

\noindent\textbf{BEV-injected Visual Features.} In the proposed GPVL, three kinds of targets in the driving scene are introduced, including foreground objects, agent motions and map elements. We first utilize the visual embedding layer to encode $\textbf{F}_{det} \in \mathbb{R}^{N_{d}\times d_{d}}$, $\textbf{F}_{motion} \in \mathbb{R}^{N_{o}\times d_{o}}$ and $\textbf{F}_{map} \in \mathbb{R}^{N_{m}\times d_{m}}$ into the same channel dimension $d_{c}$, where $N_{d}$, $N_{o}$ and $N_{m}$ represent the numbers of bounding boxes, future trajectories and map elements, while $d_{d}$, $d_{o}$ and $d_{m}$ denote the feature dimensions of boxes, motions and maps, respectively. Then, the transformer structure~\cite{vaswani2017attention} is introduced to establish compact interactions via long-range attention for enhancing visual features. The whole function can be formulated as
\begin{eqnarray}
\textbf{V}_{det}\hspace{-0.8em}&=&\hspace{-0.8em}{\rm VisAtt}({\rm Embed}(\textbf{F}_{det})), \nonumber \\
\textbf{V}_{motion}\hspace{-0.8em}&=&\hspace{-0.8em}{\rm VisAtt}({\rm Embed}(\textbf{F}_{motion})), \nonumber \\
\textbf{V}_{map}\hspace{-0.8em}&=&\hspace{-0.8em}{\rm VisAtt}({\rm Embed}(\textbf{F}_{map})),
\end{eqnarray}
where ${\rm Embed}(\cdot)$ means the feature embedding layer and ${\rm VisAtt}(\cdot)$ represents the visual attention block. Therefore, $\textbf{V}_{det}\in \mathbb{R}^{N_{d}\times d_{c}}$, $\textbf{V}_{motion}\in \mathbb{R}^{N_{o}\times d_{c}}$ and $\textbf{V}_{map}\in \mathbb{R}^{N_{m}\times d_{c}}$, where $d_{c}$ means the common channel dimension. Afterwards, the attentive detection, motion and map features are integrated along the spatial-wise dimension to obtain the global visual feature $\textbf{V}_{global}\in \mathbb{R}^{(N_{d} + N_{o} + N_{m})\times d_{c}}$ as
\begin{equation}
\textbf{V}_{global} ={\rm Cat}[\textbf{V}_{det}, \textbf{V}_{motion}, \textbf{V}_{map}],
\end{equation}
where ${\rm Cat}(\cdot)$ denotes the concatenation operation.

\noindent\textbf{Environmental Linguistic Features.} In order to endow the model with language awareness, a textual attention module is presented by utilizing the BERT structure. This module processes perception and planning labels, such as bounding boxes, future trajectories and map elements, which are transformed into driving-specific language prompts using predefined templates.
The descriptions related to detection, motion and map features are subsequently fed into the textual attention block to generate corresponding linguistic representations. Now, the entire procedure can be formulated as
\begin{eqnarray}
\textbf{T}_{det}\hspace{-1em}&=&\hspace{-1em}{\rm TxtAtt}({\rm Embed}({\rm P}^{d}[{\rm L}_{det}])), \nonumber \\
\textbf{T}_{motion}\hspace{-1em}&=&\hspace{-1em}{\rm TxtAtt}({\rm Embed}({\rm P}^{o}[{\rm L}_{motion}])), \nonumber \\
\textbf{T}_{map}\hspace{-1em}&=&\hspace{-1em}{\rm TxtAtt}({\rm Embed}({\rm P}^{m}[{\rm L}_{map}])),
\end{eqnarray}
where ${\rm L}_{det}$, ${\rm L}_{motion}$ and ${\rm L}_{map}$ represent the ground-truth labels of foreground objects, future motions and map elements, respectively. ${\rm P}^{d}$, ${\rm P}^{o}$ and ${\rm P}^{m}$ denote the template descriptions of foreground objects, future motions and map elements, respectively. ${\rm TxtAtt}(\cdot)$ represents the BERT-based textual attention block.
$\textbf{T}_{det}\in \mathbb{R}^{L_{d}\times d_{c}}$, $\textbf{T}_{motion}\in \mathbb{R}^{L_{o}\times d_{c}}$ and $\textbf{T}_{map}\in \mathbb{R}^{L_{m}\times d_{c}}$, where ${L}_{d}$, ${L}_{o}$ and ${L}_{m}$ are sentence lengths. Subsequently, these descriptive features are concatenated to form the global textual representation $\textbf{T}_{global}\in \mathbb{R}^{(\textit{L}_{d} + \textit{L}_{o} + \textit{L}_{m})\times \textit{d}_{c}}$ as
\begin{equation}
\textbf{T}_{global} ={\rm Cat}[\textbf{T}_{det}, \textbf{T}_{motion}, \textbf{T}_{map}].
\end{equation}

\noindent\textbf{Group-wise Alignment.} In order to fully exploit the multi-level associations between different features, a group-wise alignment strategy is designed by using contrastive learning~\cite{radford2021learning}. Four types of 3D-vision language~(VL) alignment groups are considered, including detection-related VL group, motion-related VL group, map-related VL group and global VL group. For example, in a batch of $K$ training samples, the alignment function aims to treat matched VL samples as positives and mismatched VL samples as negatives. Given visual feature $\textbf{X}$ and textual feature $\textbf{T}$ as inputs, the contrastive loss is defined as
\begin{equation}
\begin{split}
\mathcal{L}_{align}(\textbf{V}, \textbf{T})=-\log \frac{\exp \left(s(\textbf{V}_{i},\textbf{T}_{i}) / \tau\right)}{\sum_{j=1}^{K} \exp \left(s(\textbf{V}_{i},\textbf{T}_{j}) / \tau\right)} \\
-\log \frac{\exp \left(s(\textbf{T}_{i},\textbf{V}_{i}) / \tau\right)}{\sum_{j=1}^{K} \exp \left(s(\textbf{T}_{i},\textbf{V}_{j}) / \tau\right)},
\end{split}
\end{equation}
where $\tau$ is a learnable temperature coefficient, and $s(\cdot, \cdot)$ is a similarity function. The function $s(\cdot, \cdot)$ is obtained by computing the similarity score between visual and textual features processed by global average pooling. In this module, a learnable weight is introduced to explore fine-grained relations between different representations. For instance, the similarity function of detection-related VL group can be formulated as
\begin{equation}
\begin{split}
s(\textbf{V}_{det}, \textbf{T}_{det})=\frac{1}{2} \max _{j=1} ^{\textit{L}_{d}}(\textbf{W}_{1} \textbf{V}_{det} \textbf{T}_{det}^{j}) + \\ \frac{1}{2} \max _{j=1} ^{\textit{N}_{d}}(\textbf{W}_{2} \textbf{T}_{det} \textbf{V}_{det}^{j}),
\end{split}
\end{equation}
where $\textbf{W}_{1} \in \mathbb{R} ^ {1 \times N_{d}}$ and $\textbf{W}_{2} \in \mathbb{R} ^ {1 \times L_{d}}$. Therefore, the total group-wise alignment loss is defined as
\begin{equation}
\begin{split}
\mathcal{L}_{ga}\!= \!\mathcal{L}_{align}(\textbf{V}_{det}, \textbf{T}_{det})\! +\! \mathcal{L}_{align}(\textbf{V}_{motion}, \textbf{T}_{motion})
\\+ \mathcal{L}_{align}(\textbf{V}_{map}, \textbf{T}_{map})\!+\!\mathcal{L}_{align}(\textbf{V}_{global}, \textbf{T}_{global}).
\end{split}
\end{equation}

\subsection{Planning via Cross-modal Language Model}
\label{sect:prop:dec}

The ego-agent dynamic interaction is an essential issue in the autonomous driving system. Previous studies~\cite{hu2023planning, jiang2023vad} have attempted to introduce learnable queries to model ego-agent relations for the query feature of ego vehicle, and it is sent into a multi-layer perceptron~(MLP) to acquire expected future trajectories. Although this strategy can present favorable performance on the specific benchmark dataset, directly generating the trajectory using MLP can lead to overfitting and have difficulty in contextual relation inference among generated waypoints. Facing these challenges, a cross-modal language model for generative planning is developed to empower GPVL with the capability to make the safe decision in a rational and robust manner.

\noindent\textbf{Ego-agent Cross-modal Decoder.} The proposed model formulates a language prompt for the self-driving car's current status, which includes its high-level driving command and location. The informative prompt is sent to the text embedding layer to yield the initial linguistic representation of the self-driving car. Meanwhile, as shown in Fig.~\ref{fig:pipeline-overall}, the designed scene captioning model generates visual descriptions (\emph{e.g.}, traffic lights, signs, crucial objects and weather). Subsequently, the environmental visual feature $\textbf{V}_{global}$ and the prompt feature are fed into a language model to learn the planning feature with rich driving scenario and navigation information. The process can be formulated as
\begin{equation}
\textbf{F}={\rm LM}({\rm Embed}({\rm Cap}), {\rm Embed}({\rm Nav}), \textbf{V}_{global}),
\end{equation}
where ${\rm LM}(\cdot)$ denotes the pre-trained language model, $\rm Cap$ is the 2D scene captioning, $\rm Nav$ means the high-level navigation and its detailed template is ``The box of ego-car is [cx, cy, cz, cw, ch, cl]. It is currently \{command\} and can not collide with other vehicles or the boundaries of the BEV map. Please predict the trajectory for the next p timestamps.'' Eventually, $\textbf{F}\in \mathbb{R}^{N \times d_{c}}$ is sent into the Linear~\cite{lecun2015deep} layer and the Softmax~\cite{bishop2006pattern} layer to yield the word vector as
\begin{equation}
\label{eq:linear}
p({w}_{t} | {w}_{1:t-1})=Softmax(Linear(\textbf{F})),
\end{equation}
where $p$ means the probability prediction of vocabulary.

\noindent\textbf{Generative Planning.} Similar to most visual captioning tasks~\cite{pan2020x, li2022taking, li2022knowledge}, a cross-entropy loss is introduced to output the trajectory in a language modeling fashion. During training, each sample $\mathcal{X}$ is equipped with a prompt captioning ${\rm Gt}$ as a reference, which includes high-level command and future trajectory of ego vehicle, and the loss function is expressed as
\begin{equation}
\label{eq:caploss}
\mathcal{L}_{cap}=- \sum _{t=1} ^{T} log (p_{\theta} (w_{t} | ({\rm Gt})_ {1:t-1})),
\end{equation}
where $\theta$ represents all trainable parameters during training and $w_{t}$ means the $t$-th predicted word.

\subsection{Training Loss}

The overall loss $\mathcal{L}_{gpvl}$ of the proposed GPVL comprises three training components, including visual perception loss, 3D-vision language alignment loss and trajectory captioning loss, which can be defined as
\begin{equation}
\label{eq:totalloss}
\mathcal{L}_{gpvl}=\mathcal{L}_{vis} + \mathcal{L}_{ga} + \mathcal{L}_{cap} ,
\end{equation}
where $\mathcal{L}_{vis}$ means the losses associated with visual downstream tasks, including 3D object detection, map construction and motion prediction.

\section{Experiment}
\label{sect:exp}

\subsection{Dataset and Automatic Metric Evaluation}
\label{sect:exp:data}

\noindent\textbf{$\textbf{Dataset}$.} Substantial experiments are conducted on the complex public nuScenes~\cite{caesar2020nuscenes} dataset, which comprises 1,000 traffic scenarios, and the duration of each video is around 20 seconds. This dataset offers over 1.4 million 3D bounding boxes across 23 different object categories. The scene images are recorded by six cameras, and the key frames are labeled 2 times per second.

\noindent\textbf{$\textbf{Evaluation Metric}$.} Two widely used objective metrics are employed to comprehensively validate the planning results, including displacement error and collision rate, which are denoted as L2 and Collision for short in the following result presentation, respectively. In particular, L2 measures the distance between predicted trajectory and ground-truth trajectory, and Collision evaluates the frequency of collisions that occur in real-world driving scenarios. Additionally, the Latency and FPS metrics are introduced to evaluate the real-time performance of models.

\subsection{Implementation Details}
\label{sect:exp:impl}

The proposed model aims to predict the trajectory for the future 3 seconds. The input image size is $1280 \times 720$. GPVL utilizes ResNet50~\cite{he2016deep} to extract the multi-view image features. The numbers of BEV queries, bounding boxes and map points are $200 \times 200$, $200$ and $100 \times 20$, respectively. The feature dimension and hidden size are 768 and 512, respectively. The model utilizes the AdamW~\cite{loshchilov2017decoupled} optimizer and weight decay 0.01 in the training process. The learning rates in three training stages are $2 \times 10^{-4}$, $1 \times 10^{-4}$ and $5 \times 10^{-6}$, respectively. The BERT~\cite{devlin2018bert} structure is used by 3D-vision language pre-training and cross-modal language model. As for testing, the size of greedy search is set to 1 to generate the trajectory caption. The proposed model is trained on PyTorch framework with 8 NVIDIA RTX A6000 cards.

\begin{table*}[]
\begin{center}
\footnotesize
\begin{tabular}{l|cccc|cccc|cc}
\toprule
\multirow{2}{*}{Method} & \multicolumn{4}{c|}{L2 (m) $\downarrow$} & \multicolumn{4}{c|}{Collision (\%) $\downarrow$} & \multirow{2}{*}{Latency (ms)} & \\
& 1s & 2s & 3s & Avg. & 1s & 2s & 3s & Avg. & & \multirow{-2}*{FPS} \\ \hline
\multicolumn{11}{c}{\textbf{Non-Autoregressive}} \\ \hline
NMP$^\dagger$~\cite{zeng2019end} & - & - & 2.31 & - & - & - & 1.92 & - & - & - \\
SA-NMP$^\dagger$~\cite{zeng2019end} & - & - & 2.05 & - & - & - & 1.59 & - & - & - \\
FF$^\dagger$~\cite{hu2021safe} & 0.55 & 1.20 & 2.54 & 1.43 & 0.06 & 0.17 & 1.07 & 0.43 & - & - \\
ST-P3~\cite{hu2022st} & 1.33 & 2.11 & 2.90 & 2.11 & 0.23 & 0.62 & 1.27 & 0.71 & 628.3 & 1.6 \\
Ego-MLP~\cite{zhai2023rethinking} & 0.46 & 0.76 & 1.12 & 0.78 & 0.21 & 0.35 & 0.58 & 0.38 & - & - \\
UniAD~\cite{hu2023planning} & 0.48 & 0.96 & 1.65 & 1.03 & 0.05 & 0.17 & 0.71 & 0.31 & 555.6 & 1.8 \\
VAD~\cite{jiang2023vad} & 0.41 & 0.70 & 1.05 & 0.72 & 0.07 & 0.17 & 0.41 & 0.22 & 224.3 & 4.5 \\
BEV-Planner~\cite{li2023ego} & 0.30 & 0.52 &0.83 & 0.55 & 0.10 & 0.37 & 1.30 & 0.59 & - & - \\ \hline
\multicolumn{11}{c}{\textbf{Autoregressive}} \\ \hline
LLaVA~\cite{liu2023visual} &1.04 &1.74 &2.57 &1.79 &0.58 &1.17 &1.74 &1.16 &2532.5 & 0.4\\
Vicuna~\cite{chiang2023vicuna} &1.06 &1.80 &2.54 &1.80 &0.60 &1.21 &1.78 &1.20 &2498.9 & 0.4\\
Merlin~\cite{yu2023merlin} &1.03 &1.71 &2.40 &1.71 &0.48 &1.05 &1.77 &1.10 &- & - \\
Ominidrive~\cite{wang2024omnidrive} & 0.40 & 0.80 & 1.32 & 0.84 & \textbf{0.04} & 0.46 & 2.32 & 0.94 & - & - \\
Atlas~\cite{bai20243d} &0.52 &0.97 &1.53 &1.00 &0.15 &0.31 &0.70 &0.38 &- & - \\
\textbf{GPVL}& \textbf{0.21} & \textbf{0.39} & \textbf{0.69} &\textbf{0.43} & 0.07 & \textbf{0.09} & \textbf{0.27} & \textbf{0.14}  & 198.2& 5.1\\
\bottomrule
\end{tabular}% }
\end{center}
\caption{\textbf{Open-loop planning performance.} GPVL achieves the highest score on most evaluation metrics on the nuScenes~\cite{caesar2020nuscenes} val dataset
. LiDAR-based methods are denoted with $\dagger$. In open-loop evaluation, the ego status information of GPVL is deactivated for a fair comparison. }
\label{tab:sota-plan}
\end{table*}

\begin{figure*}[htbp]
\begin{center}
\includegraphics[width=0.68\linewidth]{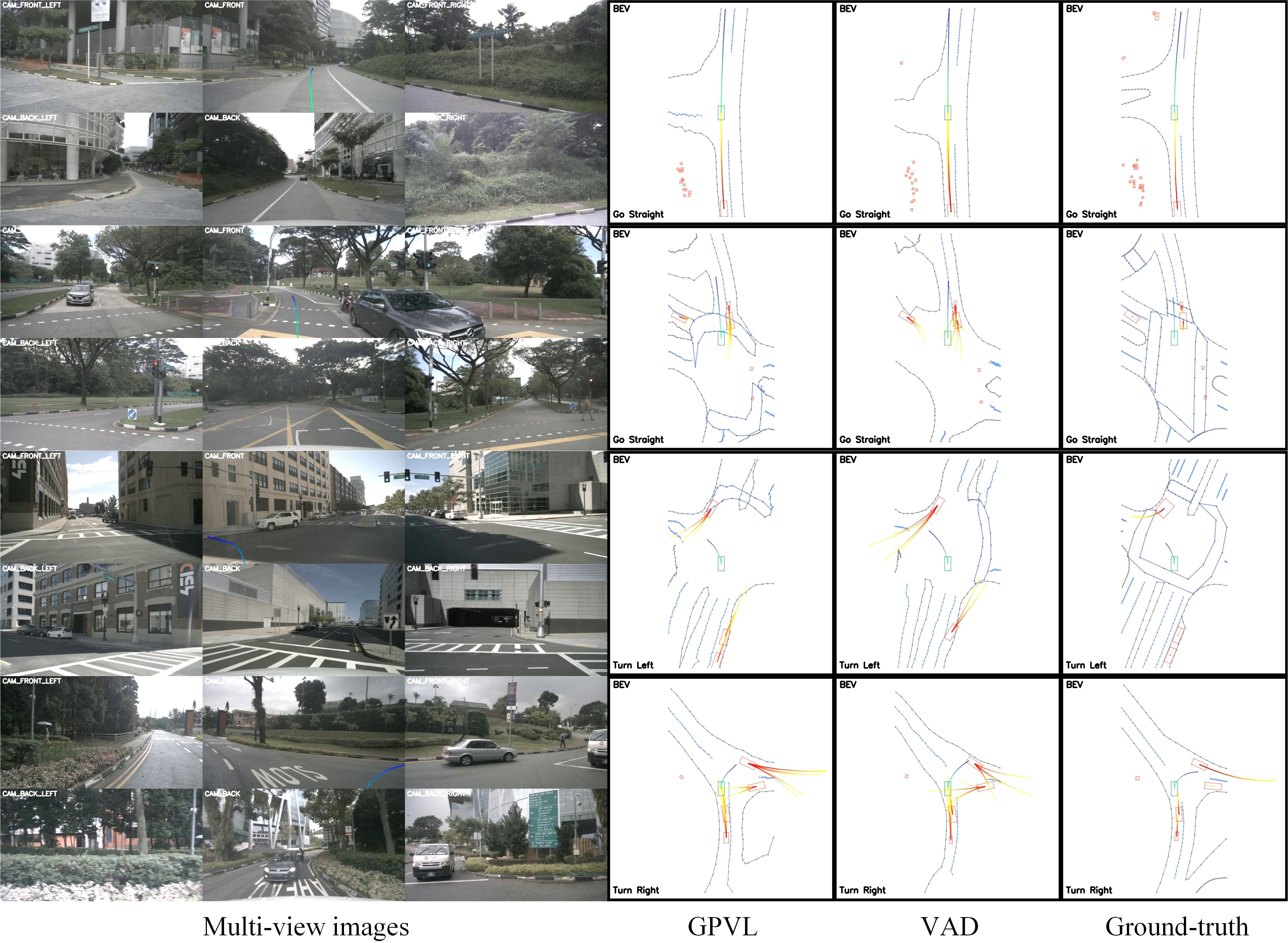}
\end{center}
\caption{Visualized comparison of the proposed GPVL, VAD and the ground-truth on the nuScenes dataset.}
\label{fig:vis}
\end{figure*}

\subsection{Comparison with State-of-the-art Methods}
\label{sect:exp:sota}

The proposed scheme is compared with 13 autonomous driving methods, including 8 traditional non-autoregressive methods and 5 LLM-injected autoregressive methods.

\subsubsection{Quantitative Result}
\label{sect:exp:sota:qnres}

Table~\ref{tab:sota-plan} shows the statistical comparisons of the proposed GPVL with other state-of-the-art methods. Generally, the statistical performance of GPVL is obviously better than other approaches. Specifically, GPVL obtains the lowest scores in L2 distance metrics and obviously reduces the planning displacement errors by 0.18m at 1s, 0.28m at 2s, 0.34m at 3s and 0.27m on average compared to VAD, indicating the superior accuracy of trajectory prediction. Moreover, GPVL exhibits the best performance in most collision rate metrics, highlighting its exceptional safety and robustness in avoiding collisions. Compared to the approaches that introduce auxiliary tasks or utilize LLMs, the proposed method maintains comparable planning performance and achieves a latency of 188.7 milliseconds and an inference speed of 5.3 fps, presenting the potential for practical applications. According to the report of BEV-Planner~\cite{li2023ego}, the proposed scheme excludes ego status information to prevent shortcut learning.

Table~\ref{tab:dif-cmds} shows the evaluated L2 and Collision scores in terms of turn left, turn right and go straight commands. It is evident that GPVL significantly outperforms UniAD and VAD across all metrics. In the nuScenes dataset, $87.7\%$ training and $88.2\%$ validation samples consist of simple go straight scenes. Therefore, UniAD and VAD are more prone to overfitting and learning shortcuts on these samples, resulting in poor performances in more complex turning scenarios. In contrast, the proposed method acquires favorable results across all scenarios, demonstrating strong generalization capabilities in diverse driving situations.

\subsubsection{Qualitative Result}
\label{sect:exp:sota:qlres}

The planning results generated by GPVL, compared to VAD and ground-truth, are illustrated in Fig.~\ref{fig:vis}. To provide a comprehensive understanding of driving scenes, multi-view camera images are included and planning trajectories are visualized in the front camera image. Generally, benefiting from the proposed 3D-vision language pre-training module and cross-modal language model, the designed model generates accurate and reasonable trajectories. For example, in the first scenario, where the ego vehicle is instructed to go straight, the trajectory generated by GPVL guides the vehicle to safely navigate through the urban road. Conversely, VAD's trajectory has a collision risk with the roadside. In the third scenario, the result of GPVL closely matches the ground-truth, enabling the vehicle to safely turn left at intersections. Nonetheless, the planning decision of VAD is more aggressive, potentially leading to collision with the oncoming vehicle.

\begin{table*}[]
\begin{center}
\renewcommand{\arraystretch}{0.9}
\renewcommand{\tabcolsep}{3.5pt}
\footnotesize
\begin{tabular}{ll|cccc|cccc|cccc}
\toprule
&\multirow{2}{*}{Methods} & \multicolumn{4}{c|}{Turn Left} & \multicolumn{4}{c|}{Turn Right} & \multicolumn{4}{c}{Go Straight} \\
& & 1s & 2s & 3s & Avg. & 1s & 2s & 3s & Avg. & 1s & 2s & 3s & Avg. \\
\midrule
\multirow{3}{*}{\begin{sideways}{L2}\end{sideways}}
&UniAD& 0.37 &0.62 &0.95 &0.65 &0.42 &0.77 &1.23 &0.80 &0.46 &0.70 &1.03 &0.73  \\
&VAD & 0.49 &0.83 &1.24 &0.85 &0.49 &0.88 &1.34 &0.90 &0.40 &0.67 &1.02 &0.70 \\
&\textbf{GPVL} & \textbf{0.29} &\textbf{0.52} &\textbf{0.88} &\textbf{0.56} &\textbf{0.27} &\textbf{0.51} &\textbf{0.85} &\textbf{0.54} &\textbf{0.19} &\textbf{0.38} &\textbf{0.66} &\textbf{0.41}  \\
\midrule
\multirow{3}{*}{\begin{sideways}{Col.}  \end{sideways}}
&UniAD & 0.15 &0.35 &0.75 &0.42 &0.09 &0.93 &1.06 &0.69 &0.10 &0.15 &0.55 &0.27  \\
&VAD & 0.13 &0.21 &0.68 &0.34 &0.18 &0.85 &1.19 &0.74 &0.12 &0.14 &0.31 &0.19 \\
&\textbf{GPVL} & \textbf{0.12} &\textbf{0.18} &\textbf{0.37} &\textbf{0.22} &\textbf{0.09} &\textbf{0.23} &\textbf{0.39} &\textbf{0.24} &\textbf{0.07} &\textbf{0.08} &\textbf{0.24} &\textbf{0.13}  \\
\bottomrule
\end{tabular}
\end{center}
\caption{Statistical results of L2 distance and collision rate~(Col.) with turn left, turn right and go straight commands.}
\label{tab:dif-cmds}
\end{table*}

\begin{table*}[htbp]
\begin{center}
\renewcommand{\arraystretch}{0.9}
\renewcommand{\tabcolsep}{4pt}
\footnotesize
\begin{tabular}{c|ccccc|cccc|cccc}
\toprule
\multirow{2}{*}{ID} & \multirow{2}{*}{Perc} & \multirow{2}{*}{Cap} & \multirow{2}{*}{VLP} & \multirow{2}{*}{GA} &  \multirow{2}{*}{CLM}  & \multicolumn{4}{c|}{L2 (m) $\downarrow$} & \multicolumn{4}{c}{Collision (\%) $\downarrow$}  \\
&  &  &  & &  & 1s & 2s & 3s & Avg. & 1s & 2s & 3s &Avg. \\
\midrule
1 & -  &\checkmark &\checkmark & \checkmark & \checkmark          & 0.99 & 1.27 &1.88 & 1.38 & 0.41 & 0.72 &1.13 & 0.75 \\
2 & \checkmark  &- & \checkmark & \checkmark & \checkmark         & 0.25 & 0.43 &0.76 & 0.48 & 0.10 & 0.12 &0.34 & 0.19 \\
3 & \checkmark  &\checkmark & - & - & \checkmark                  & 0.49 & 0.75 &1.12 & 0.79 & 0.25 & 0.33 &0.57 & 0.38 \\
4 & \checkmark  &\checkmark & \checkmark & - & \checkmark         & 0.36 & 0.59 &0.91 & 0.62 & 0.16 & 0.24 &0.48 & 0.29  \\
5 & \checkmark  &\checkmark & \checkmark &\checkmark & -          & 0.30 & 0.54 &0.85 & 0.56 & 0.11 & 0.17 &0.39 & 0.22 \\
6& \checkmark  &\checkmark & \checkmark &\checkmark & \checkmark & \textbf{0.21} & \textbf{0.39}  & \textbf{0.69} & \textbf{0.43} & \textbf{0.07} & \textbf{0.09} & \textbf{0.27}& \textbf{0.14} \\
\bottomrule
\end{tabular}
\end{center}
\caption{Ablation study of GPVL on nuScenes, where Perc, Cap, VLP, GA and CLM represent perception modules, captioning model, 3D-vision language pre-training, group-wise alignment and cross-modal language model, respectively.}
\label{tab:ablation}
\end{table*}

\subsection{Ablation Study}
\label{sect:exp:abl}

The ablation study in Table \ref{tab:ablation} systematically investigates the contributions of the key components of GPVL on the nuScenes dataset. Without the perception module, GPVL struggles with detecting foreground objects, predicting motions and constructing maps, leading to higher L2 and Collision scores. Disabling the VLP and GA components significantly degrades performance, highlighting model's strong ability to bridge visual and linguistic understanding. The absence of GA results in obvious performance degradation, indicating its importance in fine-grained feature association. The exclusion of CLM increases L2 and Collision scores, emphasizing its role in generating reasonable planning decisions. Finally, as shown in the sixth row of Table~\ref{tab:ablation}, the integration of all modules yields the best performance, showcasing the synergistic effect of the combined system.

\subsection{Zero-shot Generalization}
\label{sect:exp:zz}

\begin{table}[htbp]
\begin{center}
\renewcommand{\arraystretch}{0.9}
\renewcommand{\tabcolsep}{2.8pt}
\centering
\footnotesize
\begin{tabular}{ll|cccc|cccc}
\toprule
&\multirow{2}{*}{Methods} & \multicolumn{4}{c|}{L2 (m) $\downarrow$} &\multicolumn{4}{c}{Collision (\%) $\downarrow$}  \\
&& 1s & 2s & 3s & Avg. & 1s & 2s & 3s & Avg. \\
\midrule
\multirow{3}{*}{\begin{sideways}{Group1}\end{sideways}}
&UniAD       & 0.49 & 0.91 & 1.66 &1.02 & 0.25 & 0.44 & 0.93 &0.54 \\
&VAD                    & 0.44 & 0.78 & 1.19 &0.80 & 0.22 & 0.34 & 0.69 & 0.41 \\
&\textbf{GPVL}     & \textbf{0.28} & \textbf{0.49} & \textbf{0.81} &\textbf{0.53} & \textbf{0.11} & \textbf{0.15} & \textbf{0.32} & \textbf{0.19} \\
\midrule
\multirow{3}{*}{\begin{sideways}{Group2}\end{sideways}}
&UniAD      & 0.53 & 0.97 & 1.71 &1.07 & 0.28 & 0.41 & 0.99 & 0.56 \\
&VAD          & 0.49 & 0.72 & 1.08 &0.76 & 0.25 & 0.33 & 0.72 & 0.43 \\
&\textbf{GPVL}  & \textbf{0.31} & \textbf{0.55} & \textbf{0.84} &\textbf{0.57} & \textbf{0.09} & \textbf{0.17} & \textbf{0.33} & \textbf{0.20} \\
\bottomrule
\end{tabular}
\end{center}
\caption{To evaluate the zero-shot performance on the new city, the models are trained on Boston and tested on Singapore in Group1, and the models are trained on Singapore and tested on Boston in Group2.}
\label{tab:zero-shot}
\end{table}

\begin{table}[htbp]
\begin{center}
\renewcommand{\arraystretch}{0.8}
\renewcommand{\tabcolsep}{3.2pt}
\centering
\footnotesize
\begin{tabular}{l|cccc|cccc}
\toprule
\multirow{2}{*}{Methods} & \multicolumn{4}{c|}{L2 (m) $\downarrow$} &\multicolumn{4}{c}{Collision (\%) $\downarrow$}  \\
&  1s & 2s & 3s & Avg. & 1s & 2s & 3s & Avg. \\
\midrule
UniAD& 0.59 & 1.12 & 1.89 &1.20 & 0.21 & 0.39 & 0.92 &0.51 \\
VAD    & 0.48 & 0.77 & 1.15 &0.80 & 0.19 & 0.31 & 0.53 & 0.34 \\
\textbf{GPVL}   & \textbf{0.27} & \textbf{0.50} & \textbf{0.89} &\textbf{0.55} & \textbf{0.11} & \textbf{0.18} & \textbf{0.37} & \textbf{0.22} \\
\bottomrule
\end{tabular}
\end{center}
\caption{To validate the robustness of models in unseen scenarios, four types of noise are introduced into test images, including rain, fog, snow and darkness.}
\label{tab:noise}
\end{table}

To validate the model's generalization ability, we train and test the models on datasets constructed from two different urban environments (\emph{i.e.}, Boston and Singapore). Specifically, two groups of experiments are introduced: (1) training on Boston and testing on Singapore, (2) training on Singapore and testing on Boston.
As shown in Table~\ref{tab:zero-shot}, the evaluated scores of GPVL in both groups are obviously better than UniAD and VAD. Moreover, four kinds of noise (\emph{i.e.}, rain, fog, snow and darkness) are introduced into test images to validate the robustness of GPVL, as presented in Table~\ref{tab:noise}, these noise conditions have a significant negative impact on UniAD and VAD, while they have a minor effect on GPVL. Therefore, the outstanding performance of GPVL in various real-world scenarios demonstrates its capacity to improve the robustness and safety of autonomous driving system.

\section{Conclusion}
\label{sect:cln}

In this work, a novel generative planning with 3D-vision language model is proposed for end-to-end autonomous driving. Specifically, the 3D-vision language pre-training module is designed to integrate valuable textual information and establish a rich 3D-vision language relation, where group-wise alignment aims to exploit multi-level associations between different representations, facilitating a better understanding and reasoning of driving scenes. The cross-modal language model is developed to serve as the generative engine, which utilizes the aligned feature and navigation to produce the future trajectory in an auto-regressive manner. This generative style enables the model to make correct decisions similar to natural language modeling. The proposed GPVL constructs a unified framework that not only performs reliable planning but also exhibits superior generalization capacity across various driving scenarios. Extensive experiments on the nuScenes dataset demonstrate that GPVL significantly outperforms state-of-the-art methods. In future works, the proposed GPVL is expected to promote the development of safer and more reliable autonomous driving technology.

\section{Acknowledgments}

This work was supported in part by National Natural Science Foundation of China under Grant 62371343, and in part by Shanghai Municipal Science and Technology Major Project under Grant 2021SHZDZX0100.

\bibliography{ref}

\end{document}